\documentclass[sigconf,screen]{acmart}

\AtBeginDocument{%
  }

\usepackage[normalem]{ulem}
\usepackage{enumitem}
\usepackage{rotating}
\usepackage{hyperref}
\usepackage{booktabs}
\usepackage{balance}
\usepackage{graphicx}
\usepackage{wrapfig}
\usepackage{algorithm}
\usepackage[noend]{algpseudocode}
\usepackage{amsmath}
\usepackage{fancyhdr}
\usepackage{bbm}
\usepackage{multirow}
\usepackage{xcolor}
\usepackage{soul}
\usepackage{tikz}
\usepackage{cleveref}
\usepackage{url}
\usepackage{mathtools}
\usepackage{pifont}
\usepackage{makecell}
\usepackage{subcaption}

\usepackage{array}
\newcolumntype{L}[1]{>{\raggedright\let\newline\\\arraybackslash\hspace{0pt}}m{#1}}
\newcolumntype{C}[1]{>{\centering\let\newline\\\arraybackslash\hspace{0pt}}m{#1}}
\newcolumntype{R}[1]{>{\raggedleft\let\newline\\\arraybackslash\hspace{0pt}}m{#1}}

\setlist{noitemsep, leftmargin=*, topsep=0pt, partopsep=0pt}

\crefformat{section}{\S#2#1#3} % see manual of cleveref, section 8.2.1
\crefformat{subsection}{\S#2#1#3}
\crefformat{subsubsection}{\S#2#1#3}

\usepackage[skins]{tcolorbox}

\tcbset{
  takeaway/.style={ enhanced,width=\hsize,left=1pt,right=1pt,top=1pt,bottom=1pt,colback=black!20!orange!4!white,boxrule=1pt,colframe=black!30!orange!20!white
  },
}

\newcommand{\greendot}{\Large\textcolor{green!60!black}{\ensuremath\bullet}}

\newcommand{\SYSTEM}{EnsembleCI}
\newcommand{\CC}{CarbonCast}


\copyrightyear{2025}
\acmYear{2025}
\setcopyright{cc}
\setcctype{by}
\acmConference[E-ENERGY '25]{The 16th ACM International Conference on Future and Sustainable Energy Systems}{June 17--20, 2025}{Rotterdam, Netherlands}
\acmBooktitle{The 16th ACM International Conference on Future and Sustainable Energy Systems (E-ENERGY '25), June 17--20, 2025, Rotterdam, Netherlands}\acmDOI{10.1145/3679240.3734630}
\acmISBN{979-8-4007-1125-1/2025/06}

\begin{document}

\title{\SYSTEM{}: Ensemble Learning for Carbon Intensity Forecasting}

 \author{Leyi Yan} 
    \affiliation{ 
      \institution{University of Waterloo}
      % \city{Waterloo} 
      % \state{ON}
      %   \country{Canada}
      \country{}
    }
    % \email{l52yan@uwaterloo.ca}

 \author{Linda Wang} 
    \affiliation{ 
      \institution{University of Waterloo}
      % \city{Waterloo} 
      % \state{ON}
      %   \country{Canada}
      \country{}
    }
    % \email{lindaw@uwaterloo.ca}

 \author{ Sihang Liu} 
    \affiliation{ 
      \institution{University of Waterloo}
      % \city{Waterloo} 
      % \state{ON}
      %   \country{Canada}
      \country{}
    }
% \email{sihangliu@uwaterloo.ca}

 \author{Yi Ding} 
    \affiliation{ 
      \institution{Purdue University}
      % \city{West Lafayette} 
      % \state{IN}
      \country{}
    }
    % \email{yiding@purdue.edu}
\authornote{Corresponding faculty author: yiding@purdue.edu.}

\begin{abstract}	

Carbon intensity (CI) measures the average carbon emissions generated per unit of electricity, making it a crucial metric for quantifying and managing the environmental impact. Accurate CI predictions are vital for minimizing carbon emissions, yet the state-of-the-art method (\CC{}) falls short due to its inability to address regional variability and lack of adaptability.

To address these limitations, we introduce \SYSTEM{}, an adaptive, end-to-end ensemble learning-based approach for CI forecasting. \SYSTEM{} combines weighted predictions from multiple sublearners, offering enhanced flexibility and regional adaptability. In evaluations across 11 regional grids, \SYSTEM{} consistently surpasses \CC{}, achieving the lowest mean absolute percentage error (MAPE) in almost all grids and improving prediction accuracy by an average of 19.58\%. While performance still varies across grids due to inherent regional diversity, \SYSTEM{} reduces variability and exhibits greater robustness in long-term forecasting compared to \CC{} and identifies region-specific key features, underscoring its interpretability and practical relevance. These findings position \SYSTEM{} as a more accurate and reliable solution for CI forecasting. \SYSTEM{} source code and data used in this paper are available at \href{https://github.com/emmayly/EnsambleCI}{https://github.com/emmayly/EnsembleCI}. 

\end{abstract}

\maketitle

\section{Introduction}

% The rapid growth of computing infrastructure, particularly in datacenters and artificial intelligence (AI) systems, has significantly increased carbon emissions~\cite{gupta2022act,wu2022sustainable,acun2023carbon,faiz2024llmcarbon,shi2024greenllm,wu2025unveiling}. For example, datacenters alone account for approximately 2-4\% of global electricity consumption~\cite{lavi2022measuring}, contributing substantially to greenhouse gas emissions. AI workloads, such as large language models training and serving, demand significant energy consumption~\cite{nguyen2024towards,ding2024sustainable}, further amplifying the environmental impact. The electricity powering these systems comes from diverse sources, ranging from conventional non-renewables like coal and natural gas to renewables such as solar, wind, and hydro. Notably, non-renewable sources have considerably higher carbon emissions compared to renewables~\cite{eia-faq,world-nuclear}.  

The rapid growth of computing infrastructure, particularly in datacenters and AI systems, has significantly increased carbon emissions, with datacenters alone accounting for 4\% of global electricity consumption~\cite{gupta2022act,wu2022sustainable,acun2023carbon}. AI workloads, like training and serving large language models, further amplify this impact~\cite{faiz2024llmcarbon,nguyen2024towards,ding2024sustainable,shi2024greenllm,wu2025unveiling}. The electricity powering these systems comes from both non-renewable and renewable sources, with non-renewables having a much higher carbon emissions~\cite{eia-faq,world-nuclear}.

\emph{Carbon intensity} (CI) is a critical metric used to quantify the average carbon emissions associated with electricity generation. Expressed in gCO\textsubscript{2}e/kWh, it represents the equivalent carbon emissions per kilowatt-hour of electricity. In this paper, we adopt the generation-based definition of CI, which accounts for the average carbon emissions from power generation at each source. Accurate CI predictions are essential for reducing the operational carbon emissions of computing and energy systems through load shifting and workload scheduling~\cite{acun2023carbon,hanafy2023carbonscaler,souza2023casper}.  

% However, predicting CI is challenging due to two primary factors: regional variations and the inherent difficulty of long-term forecasting~\cite{li2024uncertainty}. Regional differences in energy sources lead to significant variability in carbon emissions, making it challenging to develop a universal model. Long-term forecasting requires capturing complex temporal patterns, which further complicates the prediction. 

% The current state-of-the-art approach, \CC{}~\cite{maji2022carboncast,maji2023multi}, represents a step forward in CI forecasting. It employs a two-tier neural network architecture and uses historical energy source production, carbon intensity, and weather data as input to predict carbon intensity up to 96 hours into the future. However, it suffers from several design limitations and inconsistent accuracy. First, \CC{} is a manually designed architecture with a fixed neural network that lacks sufficient justification for its suitability in CI prediction and fails to adapt to regional variations and energy source diversity. Second, it has an inefficient two-tier architecture that takes different inputs/outputs. Our empirical analysis (\Cref{subsec-surpass}) shows this two-tier approach offers no clear advantage over a simpler, one-tier design. 

However, predicting CI is difficult due to regional energy variability and the complexity of long-term forecasting~\cite{li2024uncertainty}, which make universal modeling and accurate temporal prediction challenging. The state-of-the-art \CC{}~\cite{maji2022carboncast,maji2023multi} approach improves CI forecasting with a two-tier neural network using historical energy, carbon, and weather data. However, it has two limitations: a fixed, manually designed architecture that does not adapt to regional or energy source differences, and an inefficient two-tier structure that, as our analysis shows, offers no clear benefit over simpler alternatives (\Cref{subsec-surpass}).

To address these limitations, we present \SYSTEM{}, an adaptive, end-to-end ensemble learning-based approach that integrates weighted predictions from a pool of predictive models (i.e.,  sublearners) for improved flexibility and accuracy, instead of relying on a single, fixed model architecture. 
This ensemble approach \emph{accommodates variations and diversity} across regional girds to achieve more accurate and reliable CI predictions, and enables an \emph{end-to-end architecture} that streamlines training and prediction by reducing redundant computations.
 
We evaluate \SYSTEM{} against \CC{} across 11 grids (6 in the US and 5 in the EU) by predicting CI based on direct emission factors up to 4 days into the future. We use \emph{mean absolute percentage error} (MAPE) as the metric. Overall, \SYSTEM{} achieves the lowest MAPE in 10–11 grids, while \CC{} only achieves the lowest MAPE in at most 1 grid. Furthermore, for day-1 through day-4 forecasts, \SYSTEM{} outperforms \CC{} by 18.1\%, 17.13\%, 19.69\%, and 23.4\%, respectively. Additionally, \SYSTEM{} demonstrates greater robustness in long-term forecasting, with smaller MAPE increases by day 4 (\SYSTEM{}: 3.63\%, \CC{}: 5.43\%). Using permutation feature importance, we identify that important features vary by grids, while sublearners strongly align in selecting the most impactful features within each grid, emphasizing their robustness. We summarize our contributions as follows.
\begin{itemize}
    \item Identifying and analyzing the limitations of the state-of-the-art CI forecasting method \CC{}.
    \item Developing \SYSTEM{}, an adaptive, end-to-end ensemble learning-based approach for accurate and flexible CI forecasting.
    \item Conducting extensive evaluation to demonstrate the effectiveness and robustness of \SYSTEM{}.
\end{itemize}
\section{Background and Motivation}

% In this section, we first introduce \CC{}, a state-of-the-art carbon intensity forecasting method, and demonstrate its performance in various grids. Next, we analyze the energy source mixes that lead to its inconsistent forecasting results. Finally, we examine alternative predictors and explore opportunities for improvement.

This section introduces \CC{}, evaluates its forecasting performance across grids, analyzes the energy mixes behind its inconsistencies, and explores alternative predictors for improvement.

\subsection{The State-of-the-Art: \CC{}}

To forecast carbon intensity using publicly available data, such as the U.S. Energy Information Administration (EIA)~\cite{eia}, without relying on third-party services like ElectricityMap~\cite{electricitymaps} and WattTime~\cite{wattime}, \CC{} has been introduced. This state-of-the-art open-source tool enables hourly carbon intensity predictions for up to 96 hours into the future~\cite{maji2022carboncast,maji2023multi}

\begin{figure}
    \centering
\includegraphics[width=1\linewidth]{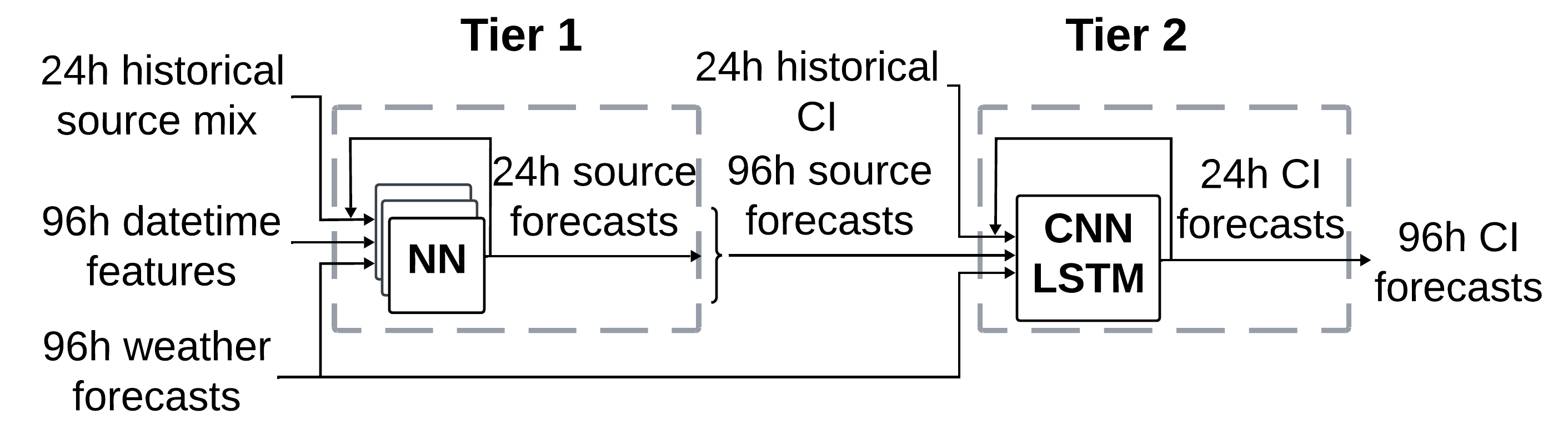}
    \vspace{-0.2in}
    \caption{The architecture design of \CC{} \cite{maji2022carboncast,maji2023multi}.}
    \label{fig:carboncast-architecture}
    \vspace{-0.15in}
\end{figure}
%% Figure source
% https://lucid.app/lucidchart/956150c7-01d2-4912-b4cc-6cc08026fe85/edit?viewport_loc=-198%2C4%2C2413%2C1199%2C0_0&invitationId=inv_955238ac-ebf2-4fcc-85d4-081126510125

% \CC{} utilizes a hierarchical two-tier architecture for carbon intensity prediction. In the first tier, an individual artificial neural network (ANN) model is employed for each energy source within a region to forecast energy production up to four days ahead. The input features for these models include historical energy production data, weather forecasts(for renewable sources), and temporal attributes such as hour-of-day and hour-of-year. The second tier integrates predicted energy production from the first tier with historical carbon intensity data and employs a CNN-LSTM model to forecast hourly carbon intensity. It uses two 1-D CNN layers to extract short-term temporal patterns and an LSTM layer to capture long-term dependencies. 

\Cref{fig:carboncast-architecture} illustrates \CC{}'s model architecture design, which employs a two-tier architecture. Tier 1 trains individual neural network models for each energy source to forecast energy production up to 4 days ahead, based on historical energy production data, weather forecasts (for renewables only), and temporal features. Tier 2 combines these energy forecasts with historical carbon intensity data, using a CNN-LSTM model with 1-D CNN layers for short-term patterns and an LSTM layer for long-term dependencies to predict hourly carbon intensity.

\begin{figure}
    \centering
    \includegraphics[width=1\linewidth]{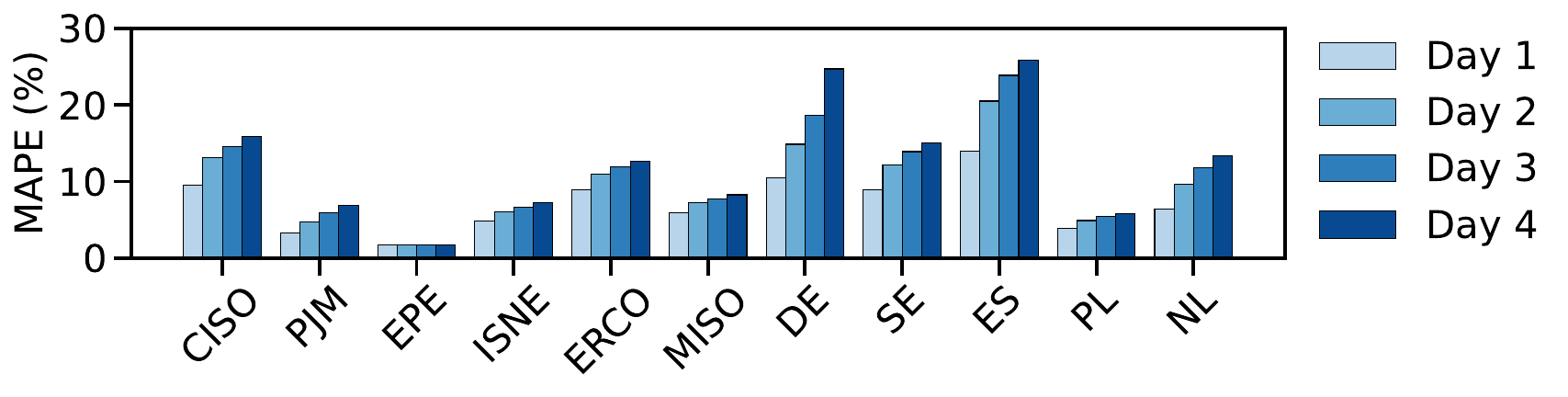}
    \vspace{-0.2in}
    \caption{Averaged hourly prediction accuracy of \CC{} for 11 regional grids over a 4-day horizon.}
    \label{fig:cc-mape}
    \vspace{-0.15in}
\end{figure}

% To evaluate \CC{}'s carbon intensity forecasting, we reproduced its training process from the original paper (detailed in \Cref{tab:regions}).
% \sihang{I recall that \CC{} does not train the whole duration of the dataset but only uses 1 year? I removed ``reproduced'' from the previous sentence.}
% expanding coverage from 6 to 11 regions (6 in the US and 5 in the EU, detailed in \Cref{tab:regions}). 
% we use \CC{} v3.0 dataset for evaluation. For 6 US regions, the model is trained using on 2019-2021 data, validated on the first half of 2022, and tested on the second half of 2022. 
% For 5 EU regions, the model is trained on 2019-2020 data, validated on the first half of 2021, and tested on the second half of 2022. 
% The model is trained on 2019–2022 data, validated on the first half of 2023, and tested on the second half. \sihang{is this correct?} \emma{US: 2019 to 2021 for testing, first half of 2022 for validation; EU: 2019 to 2020 for testing, first half of 2021 for validation}
To evaluate \CC{}'s carbon intensity forecasting, we evaluate the 11 regional grids as listed in \Cref{tab:regions} \cite{electricitymaps}. Methodology details are discussed in \Cref{sec:eval_meth}. 
% To ensure robustness, results are averaged over three stochastic runs. 
Prediction accuracy is measured using mean absolute percentage error (MAPE), where lower values indicate better performance.
\Cref{fig:cc-mape} shows the averaged hourly prediction accuracy results of \CC{} across 11 grids up to 4 days into the future. The results exhibit significant inconsistency in prediction accuracy across grids. Notably, ES recorded the highest average MAPE of 21.03\%, with values of 23.86\% and 25.81\% on the latter days. CISO (13.27\%), ERCOT (11.1\%), DE (17.18\%), NL (10.3\%), and SE (12.5\%) have moderate average MAPE values. PJM (5.21\%), EPE (1.73\%), ISNE (6.2\%), MISO (7.27\%), and PL (4.99\%) have the lowest MAPE values. This inter-regional variability in prediction accuracy raises questions about the \CC{}'s reliability and generalizability, which we will investigate next.

% \begin{table}[t]
%     \centering
%     \small
%     \caption{Electricity sources and carbon intensities in different regions in 2023.}
%     \label{tab:regions}
%     \begin{tabular}{lL{2.5cm}ll}
%     \toprule
%     Region & State/Country & Avg. CI  & Primary electricity sources \\
%     \midrule
%     CISO     &  CA    & 189 &  natural gas, solar, hydro \\
%     PJM     &  NJ    & 280 &  natural gas, nuclear, coal\\
%     EPE     &  TX, NM    &334 & natural gas, solar  \\
%     ERCOT     &  TX    & 130 & wind, coal, nuclear \\   
%     ISNE     & ME, VT, NH, MA, CT     & 241 & natural gas, nuclear, hydro\\ 
%     MISO     & MN, WI, IA, IL, IN, MI, AR, MS, LA   &361 & natural gas, coal, nuclear \\ 
%     DE     &   Germany   & 260 &  wind, coal, solar\\ 
%     SE     &  Sweden    & 117 & nuclear, natural gas, hydro\\ 
%     ES     &  Spain    & 101 & wind, nuclear, natural gas\\ 
%     NL     &   Netherlands   & 397 & natural gas, wind, coal \\ 
%     PL     &   Poland    & 527 & coal, wind, natural gas\\ 
%     \bottomrule
%     \end{tabular}\label{tbl:region}
% \end{table}

\begin{table}[t]
    \centering
    \setlength{\tabcolsep}{4pt}
    \footnotesize
    \caption{11 grids in this paper: their geographical coverage, average CI in 2024 (gCO\textsubscript{2}e/kWh), and electricity sources.}
    \vspace{-0.1in}
    \label{tab:regions}
    \begin{tabular}{lL{3.1cm}ll}
    \toprule
    Grid & State/Country   & Avg. CI & Primary electricity sources \\
    \midrule
    CISO     &  CA    & 230 &  Natural gas, solar, hydro \\
    PJM     &  IL, MI, IN, OH, PA, NJ, DE, DC, KY, WV, VA, NC  & 398   &  Natural gas, nuclear, coal\\
    EPE     &  TX, NM   &540  & Natural gas, solar  \\
    ISNE     & ME, VT, NH, MA, CT, RI  & 298  & Natural gas, nuclear, hydro\\ 
    ERCOT     &  TX  & 371  & Natural gas, wind, coal \\   
    MISO     & MN, WI, IA, IL, IN, MI, AR, MS, LA, KY, ND, TX, MO  & 485 & Natural gas, coal, wind \\ 
    DE     &   Germany  & 333  &  Wind, coal, solar\\ 
    SE     &  Sweden   & 23  & Hydro, nuclear, wind\\ 
    ES     &  Spain  & 125  & Wind, nuclear, solar\\ 
    NL     &   Netherlands & 263  & Natural gas, wind, solar \\ 
    PL     &   Poland  & 704 & Coal, wind, natural gas\\ 
    \bottomrule
    \end{tabular}
\end{table}
\vspace{-0.2in}

\begin{figure}
    \centering
    \includegraphics[width=1\linewidth]{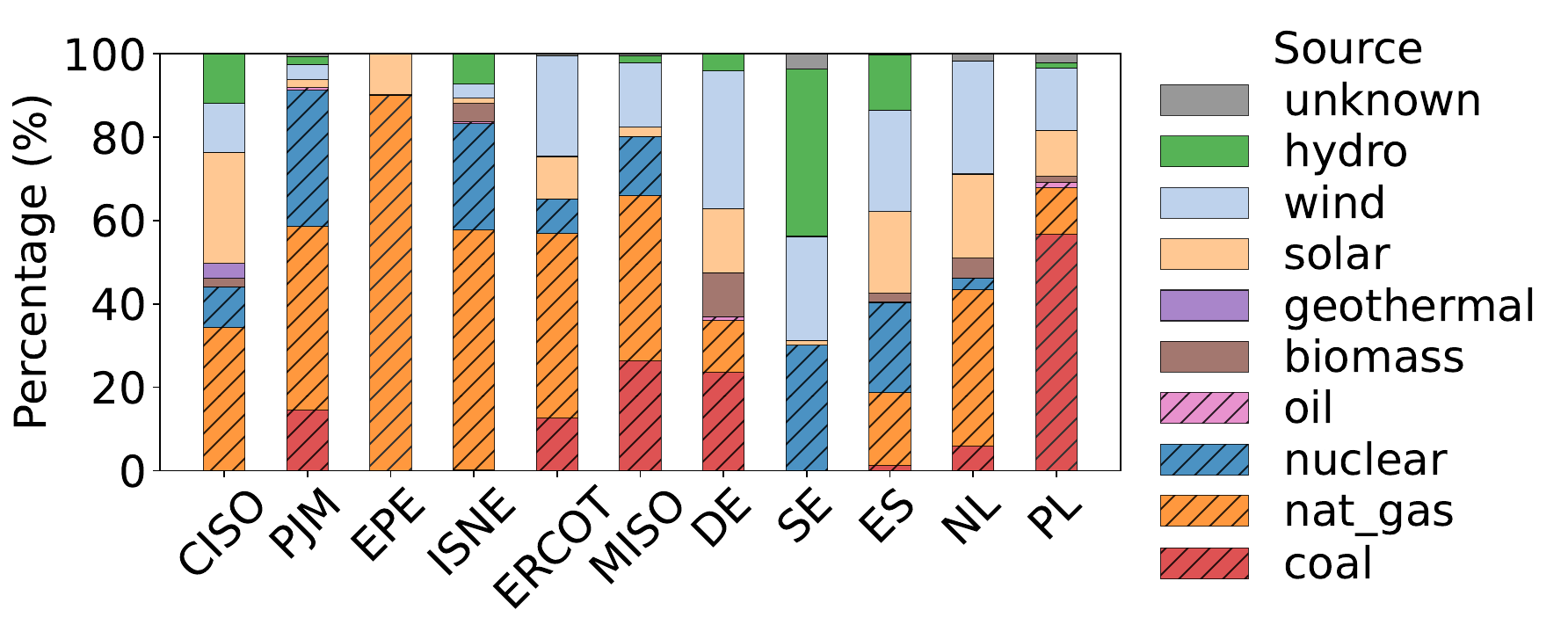}
    \vspace{-0.2in}
    \caption{Energy source mixes for 11 regional grids in 2024.}
    \label{fig:energy_sources}
    \vspace{-0.15in}
\end{figure}

\subsection{Why does \CC{} not perform well consistently across grids?}

% The dataset is from the v3.1 branch of \CC \cite{maji2022carboncast}. For each region, we include their average CI (gCO\textsubscript{2}e/kWh) in 2023 and their primary electricity sources.  

To understand why highly accurate prediction is hard and \CC{} performs inconsistently across grids, we examine the energy source mix for each grid. \Cref{fig:energy_sources} shows the breakdown of energy sources for 11 regional grids in 2024 \cite{electricitymaps}, revealing significant regional variations in both types and proportions. Analyzing these alongside prediction accuracy, we derive the following insights.

Grids with high levels of renewable energy integration, such as wind and solar, face considerable challenges in accurate forecasting due to their inherent variability. For instance, in Spain (ES), where 24.36\% of energy is derived from wind, 13.17\% from hydro, and 19.55\% from solar, the MAPE ranges from 13.93\% to 25.81\%. 
The variability in output from these sources is driven by complex environmental factors, including topography, surface roughness, temperature inversions, and cloud cover, all of which introduce significant uncertainty into predictions~\cite{solarforecast}. Conversely, grids highly relying on fossil fuels tend to achieve higher forecasting accuracy because energy production remains relatively unaffected by weather fluctuations.
In the case of EPE, where 90.12\% of energy is generated from natural gas, the MAPE of 1--4 days prediction is remarkably low, ranging between 1.68\% and 1.76\%.
%Unlike renewable sources, fossil energy production remains relatively unaffected by weather fluctuations and demonstrates stable output over time, allowing for much more precise forecasts.

% Note on edit: this is to address the peer review comment that "using an ensemble will not address this fundamental issue " and need acknowledgement

\begin{figure}
    \centering
    \includegraphics[width=1\linewidth]{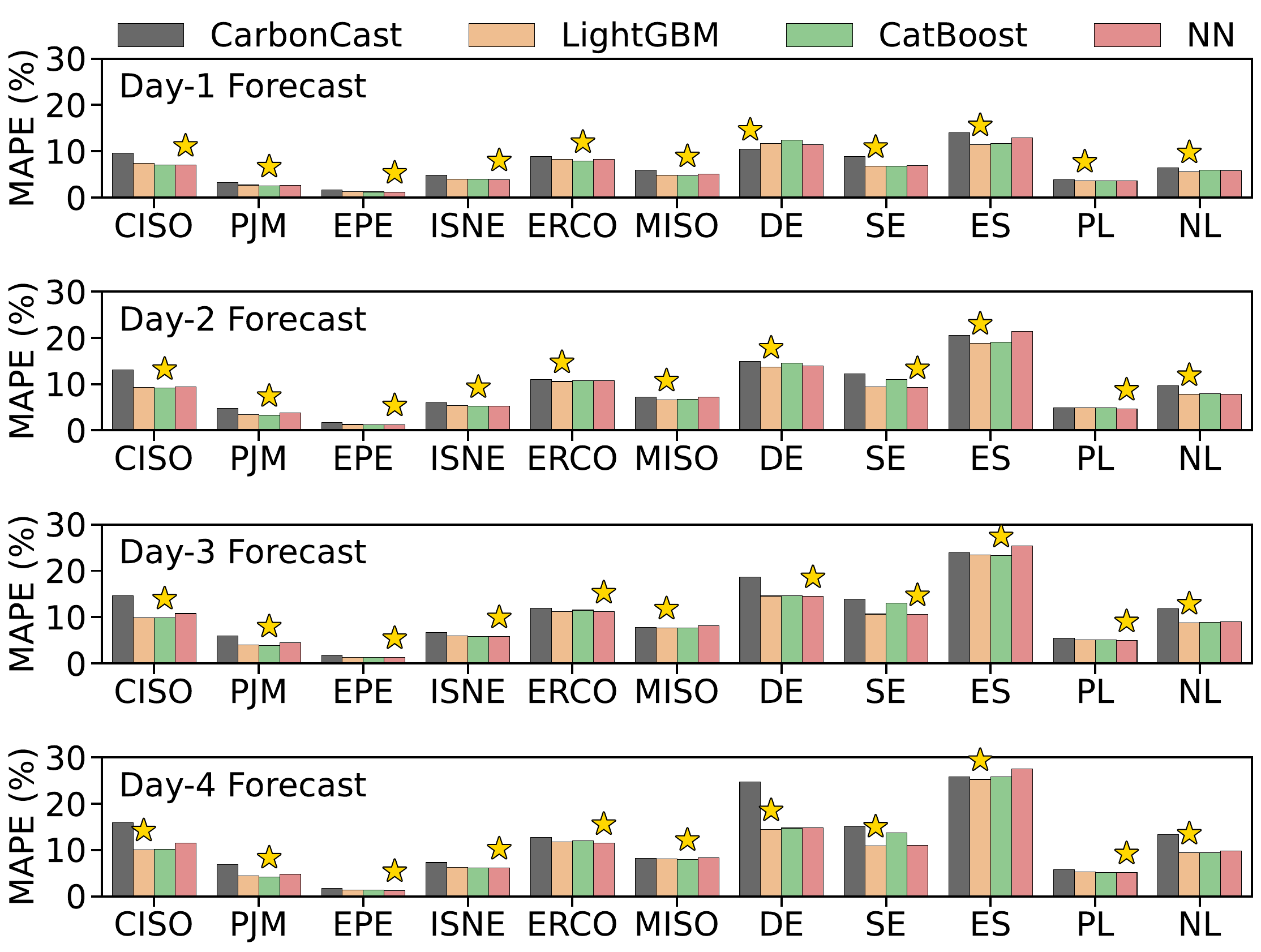}
    \caption{Prediction accuracy of different methods. The best method for each grid on each day is marked with a star.}
    \label{fig:accuracy_models}
    \vspace{-0.2in}
\end{figure}

\subsection{Can other methods surpass \CC{}?}\label{subsec-surpass}

To address \CC{}'s inconsistent performance across grids, we explore alternative models that may perform better. We hypothesize that region-specific factors require tailored modeling. To test this hypothesis, we evaluated three high-performing methods: LightGBM~\cite{NIPS2017_6449f44a}, CatBoost~\cite{dorogush2018catboost}, and one neural network model~\cite{howard2020fastai}. While we also tested models like XGBoost~\cite{chen2016xgboost} and Random Forest~\cite{breiman2001random}, these three outperformed the rest. 
% Ensemble learning, a machine learning technique that combines predictions from multiple models, is well known for outperforming individual models and reducing the variance of final predictions~\cite{dietterich2000ensemble}.

\begin{itemize}
    \item \textbf{LightGBM} is a tree-based gradient boosting framework designed for faster training, reduced memory usage, and higher accuracy~\cite{NIPS2017_6449f44a}. We select it for its outstanding performance in time series predictions, such as greenhouse temperature prediction~\cite{app13031610}.  
    \item \textbf{CatBoost} is a tree-based gradient boosting framework optimized for categorical features~\cite{dorogush2018catboost}. We select it as it automatically handles categorical features while reducing overfitting and bias~\cite{dorogush2017fighting}.  
    \item \textbf{NN} is a standard neural network implemented using the FastAI Tabular Model~\cite{howard2020fastai}. It includes an input layer followed by three fully connected layers. The first two layers apply ReLU activation, batch normalization, and dropout for regularization, with 200 and 100 neurons, respectively; the final layer outputs a single prediction value.
\end{itemize}

\Cref{fig:accuracy_models} shows that LightGBM, CatBoost, and NN outperform \CC{} in 10 grids, with average MAPE improvements of 17.57\%, 16.49\%, and 15.84\%, respectively. Their MAPE standard deviations (0.051, 0.0528, and 0.0566) are also lower than \CC{}'s (0.0574), indicating more consistent performance.

\begin{tcolorbox}[takeaway]
\textbf{Takeaways:}
\begin{enumerate}
    \item No single predictor outperforms across all grids due to each grid's unique characteristics.
    \item The tested predictors outperform mostly \CC{} in accuracy and provide more consistent results across grids.
\end{enumerate}
\end{tcolorbox}
\section{\SYSTEM{}}

Our previous analysis shows that designing a global model (i.e., \CC{} in this study) for all grids does not achieve the best prediction accuracy for each grid due to the unique characteristics of each region, such as energy source mix and weather conditions. This highlights the necessity of an adaptive approach capable of selecting highly accurate methods tailored to each grid. To address this challenge, we present \SYSTEM{}, an end-to-end ensemble learning approach that leverages a pool of highly accurate sublearners and adaptively combines them with different weights to deliver the best predictions. The \SYSTEM{} design includes two key components: subleaner selection and ensembling strategy.

\textbf{Sublearner selection.} \SYSTEM{} is a general framework that can incorporate any machine learning models as sublearners. In this paper, we select LightGBM, CatBoost, and NN evaluated in \Cref{subsec-surpass} as they have proved their ability to outperform \CC{} in most cases. Each sublearner is trained independently in the usual fashion before being combined in the ensemble. 
We do not include \CC{} as a sublearner because it has a two-tier implementation where each tier is trained independently, being unable to seamlessly integrate into the end-to-end \SYSTEM{} design. We leave a uni-tier version of \CC{} as future work. 

\begin{figure}
    \centering
    \includegraphics[width=1\linewidth]{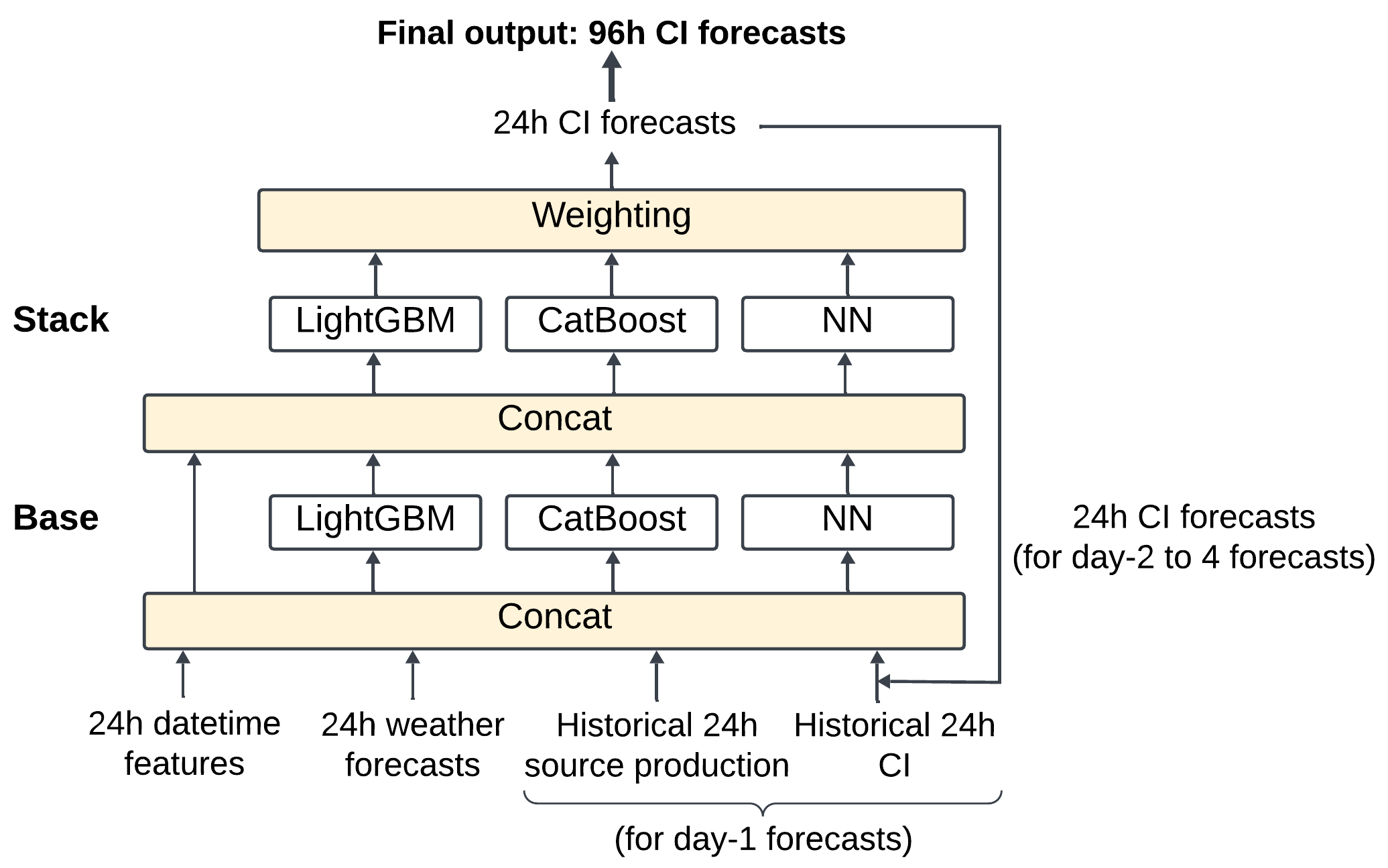}
    \caption{The architecture design of \SYSTEM{}.}
    \label{fig:architecture}
    \vspace{-0.3in}
\end{figure}
%% Figure 
% https://lucid.app/lucidchart/46008790-8d9c-4000-bf49-9c100d915019/edit?invitationId=inv_5576b226-e235-4db0-8cc4-5c18f500b54e&referringApp=slack&page=0_0#

\textbf{Ensembling strategy.} Various ensembling strategies, including bagging~\cite{breiman1996bagging}, boosting~\cite{freund1996experiments}, stacking~\cite{ting1997stacking}, and weighted combinations \cite{jimenez1998dynamically}, have been introduced. Inspired by recent AutoML frameworks like \texttt{AutoGluon}~\cite{erickson2020autogluon}, we adopt multi-layer stacking for \SYSTEM{} due to its ability to capture interactions between sublearners, enhancing final predictions~\cite{shafieian2023multi}. Stacking trains a model using the aggregated predictions of the base models as its input features.
\Cref{fig:architecture} illustrates the multi-layer stacking ensembling architecture of \SYSTEM{}, which includes two stages: base and stacking. In the base stage, \SYSTEM{} trains each sublearner individually using the raw features as input and outputs their corresponding predictions. In the stacking stage, each sublearner takes as input a concatenation of the raw features and the predictions generated by all sublearners in the base stage. To produce the final predictions, \SYSTEM{} uses ensemble selection~\cite{caruana2004ensemble} to aggregate the predictions from the stacking stage in a weighted manner.

\textbf{Model Adaptability.} The weights in \SYSTEM{} are optimized during training and remain static post-deployment. To adapt to evolving grid conditions, the model can be retrained periodically, ensuring that the weights reflect the latest energy source dynamics and weather patterns. This periodic retraining ensures that \SYSTEM{} remains adaptable to changing grid conditions, such as fluctuations in renewable energy generation, thereby improving the robustness of long-term predictions.

\textbf{Implementation.} We implement \SYSTEM{} on top of the \texttt{AutoGluon}~\cite{erickson2020autogluon} framework as it is a widely used AutoML framework to automate machine learning workflows.
We implement data preprocessing and evaluation metrics with the \texttt{scikit-learn} package \cite{pedregosa2011scikit}. As shown in \Cref{fig:architecture}, two separate models are trained for a 24-hour forecast horizon: one for the initial 24 hours (day 1) and another for the subsequent 24-hour periods (day 2–4). Each model consists of 24 sub-models, with each sub-model predicting the carbon intensity for one hour. The outputs from the Day 1 model, along with future weather forecasts and datetime features, serve as inputs to the Day 2–4 model. This model is applied recursively across three iterations to generate forecasts for the next 72 hours. Combining the outputs of both models provides a complete 96-hour carbon intensity forecast.
\section{Evaluation}

In this section, we outline our evaluation methodology, compare \SYSTEM{} and \CC{} prediction results, and discuss key features influencing the final predictions.

\subsection{Methodology}\label{sec:eval_meth}

We evaluate a total of 11 regional grids, as detailed in \Cref{tab:regions}, including 6 in the US and 5 in the EU. For a fair comparison, both \CC{} and \SYSTEM{} are evaluated using \CC{} v3.0 dataset. For US grids, training uses data from 2019 through the first half of 2022 and testing uses the second half of 2022. For EU grids, training uses data from 2019 through the first half of 2021, with testing on the second half of 2021. 
The input features include the historical 24-hour groundtruth carbon intensity values, historical 24-hour energy source production data (used only for day-1 forecasts), future 24-hour datetime features (sin/cos transformations of hour of day, day of year, and day of week), and 24-hour weather forecasts. 
As the goal is to evaluate the prediction accuracy of the models themselves, we remove the one-day source production predictions from external sources like OASIS~\cite{caiso_oasis} and ENTSO-E~\cite{entsoe_transparency} for both \CC{} and \SYSTEM{}. We use MAPE as the evaluation metric for prediction accuracy and take the average of five runs. 

\subsection{Prediction Results} \label{sec:results}
% \sihang{Emma, try to use a different rand see in \SYSTEM{} and take an avg just like \CC{}}
\Cref{tab:pred-results} shows the prediction results for \SYSTEM{} and \CC{} up to 4 days blue into the future across 11 grids. Overall, \SYSTEM{} achieves the lowest MAPE in almost all grids, except day-1 in DE.
% across 11 grids, while \SYSTEM{} only achieves the lowest at most once.
Averaged over all grids, \SYSTEM{} achieves lower MAPEs of 5.92\%, 8.16\%, 9.08\%, and 9.56\% for day-1 through day-4, respectively, compared to \CC{}'s averages of 7.05\%, 9.62\%, 11.12\%, and 12.48\%. This reflects relative improvements of 18.1\%, 17.13\%, 19.69\%, and 23.4\% across the respective horizons. This trend is evident in individual grids, such as CISO, where \SYSTEM{} achieves a MAPE of 6.67\% on day-1, reducing the error by 31.32\% compared to \CC{}’s 9.57\%. Similarly, in the PJM grid, \SYSTEM{} achieves a 38.88\% improvement on day-4, with a MAPE of 4.19\% compared to \CC{}’s 6.85\%.

Additionally, \SYSTEM{} demonstrates greater robustness for long-term forecasting. For instance, in DE, \SYSTEM{}’s MAPE increases from 11.57\% on day-1 to 14.23\% on day-4, yielding a total increase of 2.66\%. In contrast, \CC{}’s MAPE increases more sharply, rising from 10.45\% on day-1 to 24.7\% on day-4, yielding a total increase of 14.25\%, approximately five times greater than that of \SYSTEM{}. This pattern holds across other grids, where the average MAPE increase for \SYSTEM{} is 3.63\%, compared to 5.43\% for \CC{}. These results demonstrate that \SYSTEM{} is more reliable than \CC{} for long-term forecasting.

% Additionally, \SYSTEM{} demonstrates greater robustness over extended prediction horizons. For instance, in DE, \SYSTEM{}’s MAPE increases from 11.57\% on day-1 to 14.23\% on day-4, yielding an error growth rate of 0.89\%. In contrast, \CC{}’s MAPE increases more sharply, rising from 10.45\% on day-1 to 24.7\% on day-4, corresponding to an error growth rate of 4.75\%, approximately five times greater than that of \SYSTEM{}. This pattern holds across other regions, where the average error growth rate for \SYSTEM{} is 1.21\%, compared to 1.74\% for \CC{}. These results highlight \SYSTEM{}’s restrained error growth, making it a more reliable model for long-term forecasting. \emma{here i calcualte "error growth rate" for each region by $(MAPE\_day\_4 - MAPE\_day\_1)/3$, and for the average one, i average out the growth rates among regions. Not sure if its the correct term to use?}

\begin{table}[t]
    \centering
    \setlength{\tabcolsep}{2pt}
    \footnotesize
    \caption{MAPE (\%) of \SYSTEM{} and \CC{}. 
    % The last row shows the number of lowest MAPE per column.
    \vspace{-0.5em}}
    \label{tab:pred-results}
    \begin{tabular}{l|llll|llll}
    \toprule
    & \multicolumn{4}{c|}{\SYSTEM{}} &  \multicolumn{4}{c}{\CC{}~\cite{maji2022carboncast,maji2023multi}} \\ 
        Grid & day-1  & day-2 & day-3 & day-4 & day-1  & day-2 & day-3 & day-4 \\ 
    \midrule
        CISO  & \textbf{6.67} & \textbf{8.89} & \textbf{9.73} & \textbf{10.15} & 9.57 & 13.07 & 14.58 & 15.87	\\
        PJM  & \textbf{2.43} & \textbf{3.24} & \textbf{3.78} & \textbf{4.19} & 3.25 & 4.76 &5.96 & 6.85 \\
        EPE & \textbf{1.13} & \textbf{1.19} & \textbf{1.22} & \textbf{1.22} &1.68 & 1.74 & 1.75 & 1.76 \\
        ISNE  &\textbf{ 3.74} & \textbf{5.1} & \textbf{5.63} & \textbf{5.94}  &  4.8 & 6.02 & 6.69 & 7.28\\ 
        ERCOT & \textbf{7.69} & \textbf{10.3} &\textbf{10.86} & \textbf{11.23}  & 8.87 & 10.98 & 11.88 & 12.67 \\   
        MISO & \textbf{4.61} &	\textbf{6.64} &	\textbf{7.65} &	\textbf{7.91} & 5.87 &	7.2 & 7.75 & 8.26\\ 
        DE &  11.57 & \textbf{13.7} & \textbf{13.93} & \textbf{14.23} & \textbf{10.45}	&14.86 &	18.69	&24.7  \\ 
        SE  & \textbf{6.78} & \textbf{9.17} & \textbf{10.2} &	\textbf{10.37} & 8.87 & 12.2 & 13.9 &	15.03 \\ 
        ES   & \textbf{11.41} &	\textbf{19.15} &\textbf{ 23.32} & \textbf{25.55} & 13.93 & 20.05 & 23.86 & 25.81\\ 
        NL  & \textbf{5.68} &\textbf{ 7.73} & \textbf{8.68} &	\textbf{9.33} &  6.44 &	9.63 &	11.82 &	13.31 \\ 
        PL  &  \textbf{3.43} &	\textbf{4.67} & \textbf{4.87} & \textbf{4.99} &  3.87 & 4.9 &	5.39 &	5.78 \\ 
    \midrule
     Average MAPE  & \textbf{5.92} & \textbf{8.16} & \textbf{9.08} & \textbf{9.56} & 7.05 & 9.62 & 11.12 & 12.48\\
    \# Lowest MAPE  &  10 & 11 & 11 & 11 & 1 & 0 & 0 & 0\\ 
    \bottomrule
    \end{tabular}
    \vspace{-0.1in}
\end{table}
\begin{table}[t]
    \caption{Top-3 features for three grids with the highest average ranks for three grids. Green dots mark those also in the top-3 for individual sublearners.\vspace{-0.5em}}
    \label{tab:feat_imp}
    \centering
    \footnotesize\setlength{\tabcolsep}{4pt}
    \begin{tabular}{cC{0.4cm}C{0.4cm}C{0.4cm}C{0.4cm}C{0.4cm}C{0.4cm}C{0.4cm}C{0.4cm}C{0.4cm}}
        & \multicolumn{3}{c}{CISO} & \multicolumn{3}{c}{MISO} &  \multicolumn{3}{c}{DE} \\  
    \toprule
    \multicolumn{1}{r|}{LightGBM} & \greendot{}  & \greendot{} & \multicolumn{1}{c|}{} &  \greendot{} & \greendot{}&\multicolumn{1}{c|}{}   &\greendot{} &\greendot{} & \greendot{} \\
    \multicolumn{1}{r|}{CatBoost}  & \greendot{}  & \greendot{} & \multicolumn{1}{c|}{}  & \greendot{} & \greendot{} & \multicolumn{1}{c|}{\greendot{}}  & \greendot{} & \greendot{} & \greendot{} \\
    \multicolumn{1}{r|}{NN}  & \greendot{}  & \greendot{} & \multicolumn{1}{c|}{\greendot{}} & \greendot{} & \greendot{} &  \multicolumn{1}{c|}{}  &  \greendot{} & \greendot{} &   \\
    % \cmidrule{2-4}\cmidrule{6-8}\cmidrule{10-12}
    \bottomrule
    \parbox{.8cm}{Top-3\\Features} & \rotatebox[origin=c]{90}{hist\_solar} & \rotatebox[origin=c]{90}{\parbox{.7cm}{forecast\\ \quad\_dswrf}} & \rotatebox[origin=c]{90}{hist\_CI} 
         &\rotatebox[origin=c]{90}{hist\_CI} & \rotatebox[origin=c]{90}{hist\_coal} & \rotatebox[origin=c]{90}{hist\_wind} & \rotatebox[origin=c]{90}{hist\_CI} & \rotatebox[origin=c]{90}{hist\_solar} & \rotatebox[origin=c]{90}{\parbox{.7cm}{forecast\\ \quad\_dswrf}} \\
    \end{tabular}
    \vspace{-0.2in}
\end{table}

\subsection{Interpreting Features}\label{sec:feat}

We calculate feature importance for \SYSTEM{} using permutation importance~\cite{altmann2010permutation}. 
This method measures each feature's importance by assessing how prediction error changes when its values are randomly shuffled. 
We showcase three grids -- CISO, MISO, and DE -- that have unique energy sources. 
\Cref{tab:feat_imp} lists the top-3 features for each sublearner in these grids. 
% Features are ranked by importance within each sublearner, and the overall rankings are averaged across sublearners. The three features with the highest average ranks are chosen as the top three for each region. Green dots in the table highlight features also ranked in the top 3 by individual sublearners.
We observe strong alignment among sublearners in selecting the most important features, highlighting their robustness. In CISO and DE, ``hist\_solar'' and ``forecast\_dswrf'' dominate, underscoring the prevalence of solar energy and its temporal variability in these grids. Meanwhile, MISO prioritizes ``hist\_coal'' and ``hist\_wind'', reflecting its reliance on these energy sources. These results shed light on regional energy dynamics by analyzing models' feature importance.

\section{Conclusion}

We present \SYSTEM{}, an adaptive, end-to-end ensemble learning-based approach that outperforms \CC{} in both accuracy and robustness for carbon intensity forecasting. By addressing regional variability through dynamic ensemble design, \SYSTEM{} offers a scalable, reliable solution for sustainable energy management.

% \textcolor{blue}{Our approach also has some limitations that present opportunities for future work. One such limitation is the increased forecasting error in regions with higher renewable energy penetration, as accurately predicting the impact of intermittent renewable generation on carbon intensity remains challenging.}

\begin{acks}

We thank the anonymous reviewers for their valuable feedback. 
This work was supported by the Natural Sciences and Engineering Research
Council of Canada (NSERC) and the Undergraduate Research Assistantship (URA) program of the Cheriton School of Computer Science at the University of Waterloo.

\end{acks}

\bibliographystyle{plain}
\bibliography{reference}

\begin{thebibliography}{10}

\bibitem{electricitymaps}
{Electricity Maps}.
\newblock \url{https://app.electricitymaps.com/map/}.

\bibitem{wattime}
{WattTime}.
\newblock \url{https://watttime.org/}.

\bibitem{acun2023carbon}
Bilge Acun, Benjamin Lee, Fiodar Kazhamiaka, Kiwan Maeng, Udit Gupta, Manoj Chakkaravarthy, David Brooks, and Carole-Jean Wu.
\newblock Carbon explorer: a holistic framework for designing carbon aware datacenters.
\newblock In {\em Proceedings of the 28th ACM International Conference on Architectural Support for Programming Languages and Operating Systems (ASPLOS), Volume 2}, 2023.

\bibitem{eia}
{US} Energy~Information Administration.
\newblock Real-time operating grid.
\newblock \url{https://www.eia.gov/electricity/gridmonitor/dashboard/electric_overview/US48/US48}.

\bibitem{altmann2010permutation}
Andr{\'e} Altmann, Laura Tolo{\c{s}}i, Oliver Sander, and Thomas Lengauer.
\newblock Permutation importance: a corrected feature importance measure.
\newblock {\em Bioinformatics}, 2010.

\bibitem{breiman1996bagging}
Leo Breiman.
\newblock Bagging predictors.
\newblock {\em Machine learning}, 24:123--140, 1996.

\bibitem{breiman2001random}
Leo Breiman.
\newblock Random forests.
\newblock {\em Machine learning}, 45:5--32, 2001.

\bibitem{caiso_oasis}
{California Independent System Operator}.
\newblock Open access same-time information system ({OASIS}).
\newblock \url{https://oasis.caiso.com/mrioasis/logon.do}, 2025.

\bibitem{app13031610}
Qiong Cao, Yihang Wu, Jia Yang, and Jing Yin.
\newblock Greenhouse temperature prediction based on time-series features and {LightGBM}.
\newblock 2023.

\bibitem{caruana2004ensemble}
Rich Caruana, Alexandru Niculescu-Mizil, Geoff Crew, and Alex Ksikes.
\newblock Ensemble selection from libraries of models.
\newblock In {\em Proceedings of the Twenty-first International Conference on Machine Learning (ICML)}, 2004.

\bibitem{solarforecast}
D.~Chaturvedi and Isha Singh.
\newblock Solar power forecasting: A review.
\newblock 07 2016.

\bibitem{chen2016xgboost}
Tianqi Chen and Carlos Guestrin.
\newblock {XGBoost}: A scalable tree boosting system.
\newblock In {\em Proceedings of the 22nd ACM SIGKDD International Conference on Knowledge Discovery and Data Mining (KDD)}, 2016.

\bibitem{ding2024sustainable}
Yi~Ding and Tianyao Shi.
\newblock Sustainable llm serving: Environmental implications, challenges, and opportunities.
\newblock In {\em 2024 IEEE 15th International Green and Sustainable Computing Conference (IGSC)}, 2024.

\bibitem{dorogush2018catboost}
Anna~Veronika Dorogush, Vasily Ershov, and Andrey Gulin.
\newblock {CatBoost}: gradient boosting with categorical features support.
\newblock {\em arXiv preprint arXiv:1810.11363}, 2018.

\bibitem{dorogush2017fighting}
Anna~Veronika Dorogush, Andrey Gulin, Gleb Gusev, Nikita Kazeev, Liudmila~Ostroumova Prokhorenkova, and Aleksandr Vorobev.
\newblock Fighting biases with dynamic boosting.
\newblock {\em arXiv preprint arXiv:1706.09516}, 2017.

\bibitem{entsoe_transparency}
{ENTSO-E}.
\newblock {ENTSO-E} transparency platform.
\newblock \url{https://transparency.entsoe.eu/}, 2015.

\bibitem{erickson2020autogluon}
Nick Erickson, Jonas Mueller, Alexander Shirkov, Hang Zhang, Pedro Larroy, Mu~Li, and Alexander Smola.
\newblock {Autogluon-Tabular}: Robust and accurate {AutoML} for structured data.
\newblock {\em arXiv preprint arXiv:2003.06505}, 2020.

\bibitem{faiz2024llmcarbon}
Ahmad Faiz, Sotaro Kaneda, Ruhan Wang, Rita~Chukwunyere Osi, Prateek Sharma, Fan Chen, and Lei Jiang.
\newblock {LLMC}arbon: Modeling the end-to-end carbon footprint of large language models.
\newblock In {\em The Twelfth International Conference on Learning Representations (ICLR)}, 2024.

\bibitem{freund1996experiments}
Yoav Freund, Robert~E Schapire, et~al.
\newblock Experiments with a new boosting algorithm.
\newblock In {\em icml}, volume~96, pages 148--156. Citeseer, 1996.

\bibitem{gupta2022act}
Udit Gupta, Mariam Elgamal, Gage Hills, Gu-Yeon Wei, Hsien-Hsin~S Lee, David Brooks, and Carole-Jean Wu.
\newblock {ACT}: {Designing} sustainable computer systems with an architectural carbon modeling tool.
\newblock In {\em Proceedings of the 49th Annual International Symposium on Computer Architecture (ISCA)}, 2022.

\bibitem{hanafy2023carbonscaler}
Walid~A Hanafy, Qianlin Liang, Noman Bashir, David Irwin, and Prashant Shenoy.
\newblock {CarbonScaler}: Leveraging cloud workload elasticity for optimizing carbon-efficiency.
\newblock {\em Proceedings of the ACM on Measurement and Analysis of Computing Systems (POMACS)}, 2023.

\bibitem{howard2020fastai}
Jeremy Howard and Sylvain Gugger.
\newblock Fastai: a layered api for deep learning.
\newblock {\em Information}, 2020.

\bibitem{jimenez1998dynamically}
D.~Jimenez.
\newblock Dynamically weighted ensemble neural networks for classification.
\newblock In {\em IEEE International Joint Conference on Neural Networks Proceedings. IEEE World Congress on Computational Intelligence (Cat. No.98CH36227)}, volume~1, pages 753--756 vol.1, 1998.

\bibitem{NIPS2017_6449f44a}
Guolin Ke, Qi~Meng, Thomas Finley, Taifeng Wang, Wei Chen, Weidong Ma, Qiwei Ye, and Tie-Yan Liu.
\newblock {LightGBM}: A highly efficient gradient boosting decision tree.
\newblock In I.~Guyon, U.~Von Luxburg, S.~Bengio, H.~Wallach, R.~Fergus, S.~Vishwanathan, and R.~Garnett, editors, {\em Advances in Neural Information Processing Systems (NeurIPS)}, 2017.

\bibitem{li2024uncertainty}
Amy Li, Sihang Liu, and Yi~Ding.
\newblock Uncertainty-aware decarbonization for datacenters.
\newblock In {\em Workshop on Sustainable Computer Systems (HotCarbon)}, 2024.

\bibitem{maji2022carboncast}
Diptyaroop Maji, Prashant Shenoy, and Ramesh~K Sitaraman.
\newblock Multi-day forecasting of electric grid carbon intensity using machine learning.
\newblock In {\em Proceedings of the 9th ACM International Conference on Systems for Energy-Efficient Buildings, Cities, and Transportation (BuildSys)}, 2023.

\bibitem{maji2023multi}
Diptyaroop Maji, Prashant Shenoy, and Ramesh~K Sitaraman.
\newblock Multi-day forecasting of electric grid carbon intensity using machine learning.
\newblock {\em ACM SIGENERGY Energy Informatics Review}, 3(2):19--33, 2023.

\bibitem{nguyen2024towards}
Sophia Nguyen, Beihao Zhou, Yi~Ding, and Sihang Liu.
\newblock Towards sustainable large language model serving.
\newblock In {\em The 3rd Workshop on Sustainable Computer Systems (HotCarbon)}, 2024.

\bibitem{pedregosa2011scikit}
Fabian Pedregosa, Ga{\"e}l Varoquaux, Alexandre Gramfort, Vincent Michel, Bertrand Thirion, Olivier Grisel, Mathieu Blondel, Peter Prettenhofer, Ron Weiss, Vincent Dubourg, et~al.
\newblock Scikit-learn: Machine learning in {Python}.
\newblock {\em the Journal of Machine Learning Research}, 2011.

\bibitem{shafieian2023multi}
Saeed Shafieian and Mohammad Zulkernine.
\newblock Multi-layer stacking ensemble learners for low footprint network intrusion detection.
\newblock {\em Complex \& Intelligent Systems}, 2023.

\bibitem{shi2024greenllm}
Tianyao Shi, Yanran Wu, Sihang Liu, and Yi~Ding.
\newblock {GreenLLM}: Disaggregating large language model serving on heterogeneous {GPUs} for lower carbon emissions.
\newblock {\em arXiv preprint arXiv:2412.20322}, 2024.

\bibitem{souza2023casper}
Abel Souza, Shruti Jasoria, Basundhara Chakrabarty, Alexander Bridgwater, Axel Lundberg, Filip Skogh, Ahmed Ali-Eldin, David Irwin, and Prashant Shenoy.
\newblock {CASPER:} carbon-aware scheduling and provisioning for distributed web services.
\newblock In {\em Proceedings of the 14th International Green and Sustainable Computing Conference (IGSC)}, page 67–73, 2023.

\bibitem{ting1997stacking}
Kai~Ming Ting and Ian~H. Witten.
\newblock Stacking bagged and dagged models.
\newblock In {\em Proceedings of the Fourteenth International Conference on Machine Learning (ICML)}, 1997.

\bibitem{eia-faq}
{U.S. Energy Information Administration}.
\newblock How much of {U.S.} carbon dioxide emissions are associated with electricity generation?
\newblock \url{https://www.eia.gov/tools/faqs/faq.php?id=77&t=11}, 2023.

\bibitem{world-nuclear}
{World Nuclear Association}.
\newblock Carbon dioxide emissions from electricity.
\newblock \url{https://world-nuclear.org/information-library/energy-and-the-environment/carbon-dioxide-emissions-from-electricity}, 2024.

\bibitem{wu2022sustainable}
Carole-Jean Wu, Ramya Raghavendra, Udit Gupta, Bilge Acun, Newsha Ardalani, Kiwan Maeng, Gloria Chang, Fiona Aga, Jinshi Huang, Charles Bai, et~al.
\newblock Sustainable {AI}: Environmental implications, challenges and opportunities.
\newblock {\em Proceedings of Machine Learning and Systems (MLSys)}, 2022.

\bibitem{wu2025unveiling}
Yanran Wu, Inez Hua, and Yi~Ding.
\newblock Unveiling environmental impacts of large language model serving: A functional unit view.
\newblock {\em arXiv preprint arXiv:2502.11256}, 2025.

\end{thebibliography}

\end{document}